\def\year{2020}\relax
\documentclass[letterpaper]{article} % DO NOT CHANGE THIS
\usepackage{aaai20}  % DO NOT CHANGE THIS
\usepackage{times}  % DO NOT CHANGE THIS
\usepackage{helvet} % DO NOT CHANGE THIS
\usepackage{courier}  % DO NOT CHANGE THIS
\usepackage[hyphens]{url}  % DO NOT CHANGE THIS
\usepackage{graphicx} % DO NOT CHANGE THIS
\usepackage{graphicx}  % DO NOT CHANGE THIS

\usepackage{amssymb}
\usepackage{amsmath}
\usepackage{multirow}
\usepackage{booktabs}
\usepackage{adjustbox}

\title{Pattern-aware Data Augmentation for Query Rewriting in Voice Assistant Systems}
\author{Yunmo Chen\textsuperscript{\rm 1}\thanks{~The work was completed when the author was an intern at Amazon.} \quad
Sixing Lu\textsuperscript{\rm 2} \quad Fan Yang\textsuperscript{\rm 2} \quad Xiaojiang Huang\textsuperscript{\rm 2} \quad  Xing Fan\textsuperscript{\rm 2} \quad  Chenlei Guo\textsuperscript{\rm 2} \\
\Large \textsuperscript{\rm 1}Johns Hopkins University \quad \textsuperscript{\rm 2}Amazon.com, Inc. USA
}
\begin{document}

\maketitle

\begin{abstract}
  Query rewriting (QR) systems are widely used to reduce the friction caused by errors in a spoken language understanding pipeline. However, the underlying supervised models require a large number of labeled pairs, and these pairs are hard and costly to be collected. Therefore, We propose an augmentation framework that learns patterns from existing training pairs and generates rewrite candidates from rewrite labels inversely to compensate for insufficient QR training data. The proposed framework casts the augmentation problem as a sequence-to-sequence generation task and enforces the optimization process with a policy gradient technique for controllable rewarding. This approach goes beyond the traditional heuristics or rule-based augmentation methods and is not constrained to generate predefined patterns of swapping/replacing words. Our experimental results show its effectiveness compared with a fully trained QR baseline and demonstrate its potential application in boosting the QR performance on low-resource domains or locales.
\end{abstract}

\section{Introduction}
\label{sec:intro}

%   Given a query rewriting (QR) system that serves to minimize the gap between human intention and machine interpretations, a lack of labelled training pairs in a specific scenario is a well-known problem. For example, a voice assistant launched in a new locale does not have sufficient user inputs and corresponding ground truth rewrites to train a comprehensive model that corrects errors from Audio Speech Recognition (ASR). Such data scarcity problems are usually addressed using data augmentation. 
  
  % put some motivation parts from Approach section here
  
  Spoken language understanding (SLU) is widely used in voice assistant systems such as Amazon Alexa and Google Home, to extract the semantic information from an input voice query. Two main components in the SLU system, automatic speech recognition (ASR) and natural language understanding (NLU) might introduce errors from many resources such as the speaker's accent or semantics ambiguity. These errors cascade and result in user dissatisfaction. Quite a few works are focusing on friction correction on the ASR component or the NLU component, including the query rewriting (QR) system \cite{Chen2020PreTrainingFQ} that leverages a deep learning architecture to handle longer context with little feature engineering. However, supervised QR learning requires a large quantity of paired utterance strings as training labels, which are usually collected through human annotation or the user's own rephrases. Such pairs are limited for small domains or locales that lack data. Moreover, the annotation cost is significant given the high demand of annotations.
  
  \begin{figure}
      \centering
      \includegraphics[width=0.8\linewidth]{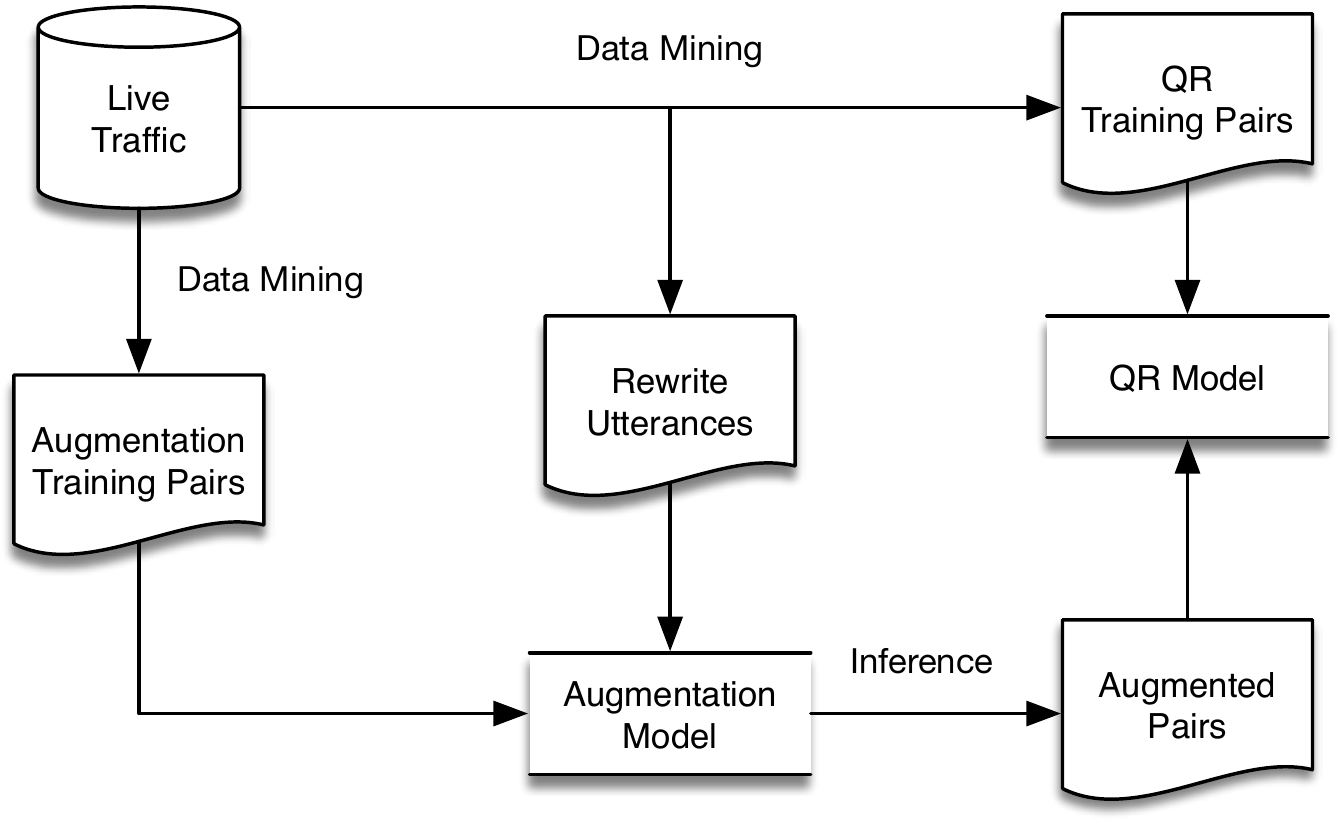}
      \caption{Augmentation System Design Overview.}
      \label{fig:ovewvier} 
  \end{figure}
  
  To conquer the data scarcity problem, data augmentation has been studied in natural language processing (NLP). Approaches spanning from deterministic text editing to neural sequence-to-sequence modeling emerged in the past few years.
  \cite[\textit{inter alia}]{WeiZ19} employed text editing techniques, where a series of candidate operations including replacement, insertion, swap, and deletion are performed to modify text sequences and introduce noise. But such pre-defined operations are limited to certain modifications and prone to generating unrealistic data examples.
  \cite[\textit{inter alia}]{SennrichHB16} showed the effectiveness of a back-translation method to augment neural machine translation with additional parallel data generated from monolingual corpora. 
  \cite[\textit{inter alia}]{CulkingHSQVD20} investigated paraphrastic augmentation which automatically expands the overall sizes and syntactic diversity via a paraphrasing model. 
%   Prior work has been 
  Although their synthetic data is meaningful in semantics, they do not model the discrepency between  friction information thus cannot model specific error patterns in the QR scenario.
  
%   However, as QR featured differently when compared to other texts, the adaptation should take its task specific signals into account (e.g. where does the . In terms of the data augmentation for QR, there are even rare prior work 

  We cast the data augmentation task as a machine translation (MT) task, augmenting training pairs by translating from existing utterances with low friction (seen as rewrites) to the potential corrupted utterances (seen as requests). This framework is an inverse process of QR that denoises requests to rewrites. 
  It follows the nature of voice assistant systems where all functionalities are usually limited and can be easily covered by existing correct utterances, while the corrupted utterance varies due to the large variety of noise sources. 

  In addition, the designed framework not only boosts the quantity of training data but also extracts specific error patterns without underlying rules. For example, requests with ASR errors tend to be similar in phonemes but different in semantic meanings. To this end, we optimize the generation model by employing a reward that measures a underlying error pattern (e.g. phonetic dissimilarity for ASR errors). The policy gradient technique is adopted to optimize such discrete rewards along with the maximum likelihood estimation.
  
  We evaluate our proposed approach on English datasets. As the approach is agnostic to a specific language, it can be easily applied to other languages or even multilingual settings.
  To summarize, our contributions are as follows:
  
  \begin{enumerate}
      \item Design a learning-based data augmentation framework that boosts training data pairs without human annotations, reducing the cost for QR system.
      
      \item Optimize the sequence generation through predefined rewards that measures distance between request and rewrite, Particularly phoneme and semantic distance in this paper, as they represent two main friction root causes in a spoken dialog system.
  \end{enumerate}

\section{Approach}
\label{sec:method}

\subsection{Problem formulation}
\label{subsec:problem}

  For QR systems, each training example can be denoted as a pair of request utterance $U = (u_1, \cdots, u_n)$ and rewrite utterance $R = (r_1, \cdots, r_m)$, where $u_i \in \mathcal{V}$ and $r_j \in \mathcal{V}$. $\mathcal{V}$ is a token vocabulary. As opposite to the QR that learns a function of $U \to R$, our data augmentation task is to learn a mapping $R \to U$. 
%   The motivation of formulating this direction is that the tasks in a intent-specific dialog system are explicit and it is easier to get the succeed utterance as ``rewrites''. 
  Under this mapping, a $U$ is guaranteed to be paired with its corresponding input $R$, which can thus be used as a training example for QR. In addition, as both $U$ and $R$ are sequences, we leverage sequence-to-sequence technique to model the generation process.
  
\subsection{Data augmentation as machine translation}

%   Our goal of data augmentation is to generate QR training examples in a low cost manner. As mentioned in the Section \ref{subsec:problem} that such examples can be denoted as a pair of two sequences $(U, R)$ and rewrites are comparably easier to obtain. It is natural to cast the augmentation problem as a translation problem, where we translate utterances that can be correctly interpreted (known as rewrites) to potentially corrupted request utterances. As the translation conditions on the input rewrites, the generated request can be easily paired with its corresponding rewrite, fulfilling the form of QR training examples.
  
%   \textbf{Translation model} The translation task here is to learn a function $f_{\theta}$, with parameters $\theta$, which maps an encoded rewrite sequence to its possible request sequence(s).
  We adopt the Transformer \cite{VaswaniSPUJGKP17} architecture to model the conditional generation process and take a rewrite utterance $R$ as input. We first encode the  utterance using the Transformer encoder so that each token $r_j$ has one vector representation as: 
  \begin{equation}
      \mathbf{r}_j = \text{Emb}(r_j) \in \mathbb{R}^{d_{tok}}
  \end{equation}
  Therefore, the rewrite utterance becomes a sequence of vectors $\mathbf{R} = (\mathbf{r}_1, \mathbf{r}_2, \cdots, \mathbf{r}_m)$. A Transformer decoder takes the encoded rewrite sequence and autoregressively computes hidden states denoted as $\mathbf{h}_t \in \mathbb{R}^{d_{hid}}$. For each hidden state, we pass the vector $\mathbf{h}_t$ to a linear layer with the output size being the size of the vocabulary $\mathcal{V}$. Softmax is applied to the output of size $|\mathcal{V}|$, yielding a distribution over the set of tokens:
   \begin{equation} \label{eq:decode}
        p_{\theta}(u_{t}|u_1,\cdots,u_{t-1}) = \frac{\exp \mathbf{w}_u^{\rm T}{\mathbf{h}_t}} {{\displaystyle\sum}_{u^\prime \in V(e)}~\exp \mathbf{w}_{u^\prime}^{\rm T}{\mathbf{h}_t}}
  \end{equation}
  The log probability of the target sequence can thus be written as:
  \begin{align}
    \begin{split}
      \log p_{\theta}(u_1, u_2, \cdots, u_T) &= \log \displaystyle \prod_{t=1}^{T} p_{\theta}(u_t|u_1, \cdots, u_{t-1}) \\
      &= \displaystyle \sum_{t=1}^{T} \log p_{\theta}(u_t|u_1, \cdots, u_{t-1})
    \end{split}
  \end{align}
  Cross-entropy loss is used to minimize the negative log probability of the target sequence given the model parameters:
  \begin{align}
    \begin{split}
      L(\theta) &= - \log p_{\theta} (\mathbf{U})\\
      &= - \log p_{\theta}(u_1, u_2, \cdots, u_T)
    \end{split}
  \end{align}
  During the training, we incorporate the teacher forcing algorithm \cite{WilliamsZ89} to force the model to be exposed to the ground truth tokens. 
  At the inference stage, we employ beam search to reduce search errors caused by the discrepancy between training and inference.
  
%   \textbf{Leveraging on pretrained models} Training a Transformer-based model from scratch is known to be difficult. We leverage on a pretrained Transformer-based models to get a better initialization which benefits the convergence on our learning task. This would mitigate the effort during the training that makes the model learn to encode basic semantic information. In this paper, our designated pretrained model is BART\cite{LewisLGGMLSZ20}, which is pretrained under a series of denoising tasks and particularly effective to be fine-tuned for sequence generation. 

\subsection{Pattern-aware sequence generation}

%   Using machine translation task with cross-entropy loss, our data augmentation model is trained in a discriminative fashion to maximize the probability of the observed target (request) sequence given the source (rewrite) sequence. Under this formulation, we let the model to discover and learn error patterns hidden in the training data freely without any external constraints. However, as the training data itself might be noisy, such learning strategy is greatly bound by the quality of the data and cannot address any particular patterns. To narrow this gap, we propose to design metrics that can measure specific error patterns and optimize model parameters accordingly to allow generating sequences awaring the patterns that we would like to address. As such metrics are generally non-differentiable, we use the REINFORCE algorithm \cite{Williams92} for optimization. 
  
%  \textbf{Sequence generation with REINFORCE} 
%   In order to enforce corrupted patterns to be learned by the augmentation model, we introduce a reinforcement learning (RL) setting. The generation model can be viewed as an ''agent'' to interact with an external ''environment'' (words within the utterance). The ''agent'' performs a series of ''actions'' according to a policy controlled by the parameters $\theta$ to generate a sequence. 
  To introduce pattern-aware objectives into the learning task, we employ policy gradient algorithm to approximate gradients with respect to non-differentiable metrics. Upon the end of the sequence generation process, encountering with \texttt{[EOS]} (end of sequence) token, the designated reward will computed,
%   the ''environment'' provides a reward 
  denoted as $r$. We compute the reward in the sequence-level by comparing the pair of target sequence(s) and source sequence. Under this formulation, the optimization objective is to minimize the negative expected reward:
  \begin{equation}
      L(\theta) = - \mathbb{E}_{\mathbf{U} \sim p_{\theta}} [r(\mathbf{U}, \mathbf{R})]
  \end{equation}

 % \textbf{Self-critic sequence training (SCST)}
  We incorporate the self-critic sequence training (SCST) \cite{RennieMMRG17} to approximate the gradient $\nabla_{\theta} L(\theta)$. We also set a reward baseline that is independent from the generated sequence to stabilize the policy gradient approximation. The reward obtained by the current model used at test time is set as a baseline $r(\hat{\mathbf{U}})$. The gradient is then approximated as the relative reward value against the baseline:
  \begin{equation}
    \begin{split}
      \nabla_{\theta}L(\theta) = - \mathbb{E}_{\mathbf{U} \sim p_{\theta}} 
      [(r(\mathbf{U}, \mathbf{R}) - r(\hat{\mathbf{U}}, \mathbf{R})) \nabla_{\theta} \log p_{\theta} (\mathbf{U})]
    \end{split}
  \end{equation}
  
%   to enable models to learn patterns that are biased towards a certain reward function. The algorithm enables to optimize indifferentiable metrics with respect to the model parameters. In this work, we design the metrics that reflect specific error patterns and use SCST to optimize such metrics so that models learn to generate sequences 

\subsection{Reward design}

  We design the following three reward functions. All $r$ are normalized within $r \in  [0,1]$.
  
  \textbf{Phonetic reward} This reward is to learn the pattern of error cascaded from the ASR system. To compute the reward $r_p$, we use an out-of-box grapheme-to-phoneme (G2P) model \cite{ParkK19} to convert the decoded target sequence $\mathbf{U}$ and the source  sequence $\mathbf{R}$ to their corresponding phoneme sequences, $\mathbf{U}_p$ and $\mathbf{R}_p$ respectively. We compute the normalized Levenshtein distance of the phoneme sequences as:
  \begin{align}
    \begin{split}
      r_p (\mathbf{U}, \mathbf{R}) &= \texttt{LevDist} (\texttt{G2P} (\mathbf{U}), \texttt{G2P} (\mathbf{R})) \\
      &= \texttt{LevDist} (\mathbf{U}_p, \mathbf{R}_p)
    \end{split}
  \end{align}
  
  \textbf{Utterance dissimilarity reward} We adopt a pre-trained semantic encoder that outputs a probability of the similarity between two utterances.\footnote{Note that any other semantic models/functions can substitute it.} This encoder embeds the request utterance and rewrite utterance as two vectors, $\mathbf{u}_d, \mathbf{r}_d \in \mathbb{R}^d$ respectively and uses cosine distance to measure the similarity between these two vectors. The dissimilarity reward is thus denoted as:
  \begin{align}
    \begin{split}
      r_d (\mathbf{U}, \mathbf{R}) &= 1 - \cos (\mathbf{u}_d, \mathbf{r}_d)
    \end{split}
  \end{align}
  
  \textbf{Combined reward} We also investigate a linear combination of the phonetic reward and utterance dissimilarity reward. This design is motivated by the error patterns that are similar in phonemes but different in semantics. The combination is controlled by a factor $\alpha \in [0, 1]$. The combined reward is:
  \begin{align}
      \begin{split}
          r_c (\mathbf{U}, \mathbf{R}) = \alpha r_p + (1 - \alpha) r_d \\ 
      \end{split}
  \end{align}

\section{Experiments and Discussions}
\label{sec:exps}

\subsection{Training data for augmentation model}  
  The training set for the augmentation model are anonymized rewrite-request pairs. These pairs are constructed under a data mining process using ASR n\textsuperscript{th} best transcripts from hundreds of domains. In such pairs, the best ASR string from ASR model  serves as the ``rewrite''; ``requests'' are corrupted wrong ASR output strings of the same user utterance. The statistics of this dataset are shown in Table \ref{tab:datasets} named as \textit{Augmentation} dataset. This dataset is in the form of $(\mathbf{U}^*, \mathbf{R})$ tuple, where the $\mathbf{U}^*$ is treated as a ground truth sequence that the model learns from. 
%   Samples of training data are shown in Table \ref{tab:aug-train-examples}.
%   For example, 
%   \textit{add salsa to my shopping list $\rightarrow$ ask salsa to my shopping list}, \textit{add to my wife $\rightarrow$ add to my library, air conditioning in the summer $\rightarrow$ conditioning in the summer},  and \textit{can you hear me $\rightarrow$ can you pair me}.

  % show some examples?
%   \begin{table}[t]
%     \centering
%     \adjustbox{max width=\linewidth}{
%     \begin{tabular}{l|l}
%       \toprule
%         \bf Rewrite & \bf Request \\ \midrule
%         \textit{add salsa to my shopping list} & \textit{ask salsa to my shopping list} \\
%         \textit{add to my wife} & \textit{add to my library} \\
%         \textit{air conditioning in the summer} & \textit{conditioning in the summer} \\
%         \textit{can you hear me} & \textit{can you pair me} \\
%       \bottomrule
%     \end{tabular}
%     }
%     \caption{Samples of augmentation model training data.}
%     \label{tab:aug-train-examples}
%   \end{table}
  
  At the inference and augmentation stage, the input rewrites are utterances that successfully processed by the Amazon Alexa and confirmed with positive user feedbacks. Those utterances are considered to be ``golden'' rewrite utterances from the history. Note that the ``golden'' inputs do not have the scarcity problem so that the augmentation model can produce as many output pairs as needed. For fair comparison, we only adopt rewrites in the evaluation QR system training set as the ``golden'' inputs. The statistics of the data are also shown in Table \ref{tab:datasets} named as \textit{InfRewrite} dataset.
  \begin{table}[t]
    \centering
      \adjustbox{max width=\linewidth}{
        \begin{tabular}{llr}
          \toprule
            Dataset & Split & \# of examples\\
          \midrule
            \multirow{3}{*}{Augmentation} & train & 3,471,063 \\
            & dev & 433,377 \\
            & test & 434,233 \\
            InfRewrite* & & 3,844,430 \\
          \bottomrule
        \end{tabular}
      }
    \caption{Augmentation dataset statistics. * denotes the inference data that will be used to generate requests to be paired as training data for QR system training.}
    \label{tab:datasets}
  \end{table}
  
  % test set ????
  The test data in the paper includes two sets: one is sampled from friction traffics to specifically measure ASR error recovery, and the other adopts users own rephrases\footnote{The pairs of one user's first utterance and following rephrased utterances are also annotated by human judges.} as labels. Both of them are very challenging testsets with around 10K samples for the evaluation of QR performance. 

  \subsection{Evaluation of QR retrieval system}
  
  %Evaluating the data augmentation is difficult and tricky. The result of data augmentation can only be reasonably assessed by training a target model using the augmented data. To save heavy training time, 
  
  The underlying end-to-end evaluation system is a dense retrieval model powered by faiss\cite{JohnsonDH17}\footnote{The retrieval model is only one of many models used by the whole QR system and does not reflect any actual production model or performance of the Alexa product.}. The search system indexes among user satisfied utterances from the live traffic in one month. 
  We augment the training set with synthetic data generated by our augmentation model. We then train a deep neural network on it. The training objective is to minimize the distance of the input pair and maximize the distances to its neighbor pairs.
%   The synthetic training data we augmented will be used to train the embedding vector of a deep neural network which seeks to minimize the distance of the input pair and maximize the distance to its neighbor pairs. 
  We use the precision within top rank K position (P@K) as the metric.
%   For example, when P@5=0.5, it indicates among the top 5 rewrites (the distance of these 5 rewrites' embedding has least distance with request's embedding) retrieval by search system, there is at least one matches the labelled rewrite.
  
  \subsubsection{Baselines}
  
  The baseline uses the \textit{train} split in the \textit{Augmentation} dataset only, without any synthetic data generated by the augmentation model. The baseline indicates how the end-to-end QR model performs with limited existing training pairs.
  
  \subsubsection{Training}
  
  We trained four augmentation models: 1) a finetuned BART translation model \cite{LewisLGGMLSZ20} without using any rewards; 2) a Policy Gradient based (PG-based) sequence-to-sequence model using the phonetic reward; 3) a PG-based sequence-to-sequence model using the utterance dissimilarity reward; 4) a PG-based sequence-to-sequence model using the combined reward. 

  We use a maximum number of tokens at 1024 to construct a mini-batch and set the maximum utterance length to be 25 tokens. The learning rate is set to $3  \times 10^{-5}$ and warm up for 200 steps. The total training step is set to 20,000 with early stopping. For \textit{combined reward} experiment, the weight between two rewards is set to be $\alpha = 0.5$. We use 7 AWS EC2 instances with a total of 56 Tesla V100 GPUs for training. All experiments are done using fairseq\cite{OttEBFGNGA19}.

\subsection{Results and analysis}

  \begin{table}[t]
    \centering
      \adjustbox{max width=\linewidth}{
        \begin{tabular}{c|ll}
          \toprule
            Reward & Rewrite (input) & Request (output) \\
          \midrule
            \multirow{3}{*}{Phonetic} & \textit{what is your favorite clothes} & \textit{what is your favorite close} \\
            & \textit{play love by wale} & \textit{play law by wale} \\
            & \textit{what does a dog sound like} & \textit{what does a dug sound like} \\ 
            \midrule
            \multirow{3}{*}{Semantic} & \textit{when does summer end in this year} & \textit{when does summer this year} \\
            & \textit{one hundred seventeen times two} & \textit{one hundred and seventy times two} \\
            & \textit{air conditioning in the summer} & \textit{conditioning in the summer} \\ 
            \midrule
            \multirow{3}{*}{Combined} & \textit{is today the last day of ramadan} & \textit{is today the last of ramadan} \\
            & \textit{turn your own volume down} & \textit{turn on your own volume down} \\
            & \textit{i do not appreciate you} & \textit{i don't not appreciate you} \\ 
          \bottomrule
        \end{tabular}
      }
    \caption{Sample inputs and outputs from  the augmentation models trained with different rewards.}
    \label{tab:train-examples}
  \end{table}

  \begin{table}[t]
    \centering
      \adjustbox{max width=\linewidth}{
        \begin{tabular}{lcccc}
          \toprule
            Augment Model & QR training input & Data Size & p@1 & p@5\\
          \midrule
            Baseline & Training data only & 3.45M & 0\% & 0\% (range of 50\%) \\
            Translation model & Training+Synthetic & 7.28M & 3.7\% & 2.1\% \\
            % \multirow{2}{*}{phonetic reward} 
            phonetic reward & Training+Synthetic & 7.35M & 6.5\% & 2.4\% \\
            % \multirow{2}{*}{semantic reward} 
            semantic reward & Training+Synthetic & 7.34M & 5.0\% & 2.5\% \\
            % \multirow{2}{*}{combined reward} 
            combined reward & Training+Synthetic & 7.35M & 3.7\% & 0.8\% \\
            
          \bottomrule
        \end{tabular}
      }
    \caption{Experimental results on the friction testset. precision is relative number. ``QR training input'' indicates what is the training data for the downstream QR model: \textit{Training only} only uses the \textit{train} split of the \textit{Augmentation} dataset; \textit{Training+Synthetic} combines the synthetic data and the \textit{train} split. ``Size'' indicates the resulted numbers of training examples. }
    \label{tab:resutls1}
  \end{table}
  
  \begin{table}[t]
    \centering
      \adjustbox{max width=\linewidth}{
        \begin{tabular}{lcccc}
          \toprule
            Augment Model & QR training input & Data Size & p@1 & p@5\\
          \midrule
            Baseline & Training only & 3.45M & 0\% & 0\% (range of 50\%) \\
            % \multirow{2}{*}{phonetic reward} 
            phonetic reward & Training+Synthetic & 7.35M & 0\% & 0.1\% \\
            % \multirow{2}{*}{semantic reward} 
            semantic reward & Training+Synthetic & 7.34M & 3.3\% & 1.3\% \\
            % \multirow{2}{*}{combined reward} 
            combined reward & Training+Synthetic & 7.35M & 2.5\% & 3.4\% \\
            
          \bottomrule
        \end{tabular}
      }
    \caption{Experimental results on the user rephrase testset. precision is relative number.}
    \label{tab:resutls2}
  \end{table}

  Table \ref{tab:train-examples} shows sample inputs and outputs from the trained augmentation models. The experimental evaluation results are summarized on Table \ref{tab:resutls1} and Table \ref{tab:resutls2}\footnote{Due to business reasons, relative improvements are indicated in both tables. All numbers are statistically significant. The baseline number for Table \ref{tab:resutls1} and \ref{tab:resutls2} is in lower 50s.}.
  % the following disclaimer should be added after acceptance
%   \footnote{\textit{Disclaimer} The numbers shown in both experimental result tables do \textbf{not} relate to the Alexa production performance, and readers shall not infer the Alexa general performance based on those numbers.}. 
  There are five categories of training data for QR evaluation system: 1) baseline QR precision trained with no synthetic data; 2) QR precision trained with the merged synthetic data from the BART augmentation model; 3) QR precision trained with the merged data from the PG-based phonetic reward augmentation model; 4) QR precision trained with the merged data from PG-based utterance dissimilarity reward augmentation model; 5) QR precision trained with the merged data from the PG-based combined reward augmentation model.
  
  % pull out the synthetic rows to form another table?
  
%   The absolute precision numbers for all models are relatively low as we only evaluate with one of the QR deep models in the recall layer. 
  The complete QR system constitutes of many more precision layers to serve customers. 
  We only select one of the retrieval models to serve the purpose of proof-of-concept, hence relatively low baseline numbers for precision.
%   However, evaluating with the complete pipeline is too costly, so we only select one of the retrieval model to prove the concept.
  
  Comparing the ``Translation model'' with the baseline in Table \ref{tab:resutls1},
  there is 3.7\% improvement in P@1 and 2.1\% improvement in P@5, which demonstrate the power of the synthetic data added. Noting that the augmentation model input is the same as the original QR training pairs, it only boosts the original label's corrupted patterns. In practice, any ``golden'' utterances can be used as inputs and they will largely increase the potential pairs out of limited scopes. 
  
  Among the three PG-based augmentation models, the phonetic reward augmentation model performs the best. It improves 6.5\% P@1 and 2.4\% P@5 using the merged data. The main reason is that the first testset emphasizes on evaluating ASR friction cases and maximizing phonetic distance represents the ASR error best. Minor difference on the phonetic string may not generate corrupted utterances that have a large intention gap. 
  
  The user rephrase testest also demonstrates the power of synthetic supplements as shown in Table \ref{tab:resutls2}.
  The rephrase set emphasizes rewrites that are more likely being semantic rephrasing and intention switching. The semantic reward augmentation model achieves better performance on this testset.
%   Because this testset is the rephrase set in which the rewrite label is more on semantic rephrasing and possibly intention switches, the semantic reward augmentation model achieves better performance than phonetic reward augmentation model. 
  The merged training data increases the precision P@1 by 3.3\% and P@5 by 1.3\%. One may notice that the precision improvement is not huge. This is because our exploration for such augmentation models is on English whose original training dataset might reach the capacity of the simple evaluation model. However, even for such almost fully trained QR models, additional synthetic training data still proves to catch additional friction patterns under different and challenging evaluation sets. We believe this augmentation framework will benefit more for locales and domains with less resource, and we also would like to address such issues in the future work.
  
  % show several examples of our generated pairs?

\section{Conclusion and Future work}
\label{sec:conclusion}

  This work proposes a data augmentation framework targeting at boosting high-quality training data for query rewriting systems. The experimental results show the effectiveness of our approach. In the future, we will explore different problem formulations such as text refinement in the augmentation process, bringing in more utterance variations. Different reward designs are another promising direction to explore. Moreover, low-resource settings in domains or locales are also worth investigation.

\bibliographystyle{aaai}
\bibliography{refs}

\end{document}